\documentclass[twoside]{article}

\usepackage[accepted]{aistats2020}
\usepackage[round]{natbib}

\usepackage[svgnames]{xcolor}

\usepackage{amsthm}
\usepackage{amsfonts}
\usepackage{multirow}
\usepackage{booktabs}
\usepackage{url}
\usepackage{tikz}
\usetikzlibrary{calc,matrix}

\usepackage{float}
\DeclareMathOperator*{\argmax}{arg\,max}

\newcommand{\eg}{e.\,g., }
\newcommand{\ie}{i.\,e., }

\newtheorem{theorem}{Theorem}
\newtheorem{proposition}{Proposition}

\DeclareMathOperator{\Tr}{Tr}

\begin{document}



\twocolumn[

\aistatstitle{Rep the Set: Neural Networks for Learning Set Representations}

\aistatsauthor{Konstantinos Skianis \And Giannis Nikolentzos \And  Stratis Limnios \And Michalis Vazirgiannis}

\aistatsaddress{\'Ecole Polytechnique \And \'Ecole Polytechnique \& \\ AUEB \And \'Ecole Polytechnique \And  \'Ecole Polytechnique \& \\ AUEB} ]

\begin{abstract}
In several domains, data objects can be decomposed into sets of simpler objects.
It is then natural to represent each object as the set of its components or parts.
Many conventional machine learning algorithms are unable to process this kind of representations, since sets may vary in cardinality and elements lack a meaningful ordering.
In this paper, we present a new neural network architecture, called RepSet, that can handle examples that are represented as sets of vectors.
The proposed model computes the correspondences between an input set and some hidden sets by solving a series of network flow problems.
This representation is then fed to a standard neural network architecture to produce the output.
The architecture allows end-to-end gradient-based learning.
We demonstrate RepSet on classification tasks, including text categorization, and graph classification, and we show that the proposed neural network achieves performance better or comparable to state-of-the-art algorithms.
\end{abstract}

\section{Introduction}\label{introduction}
In a variety of domains, complex data objects can be expressed as compositions of other, simpler objects.
These simpler objects naturally correspond to the parts or components of the complex objects.
For instance, in natural language processing, documents may be represented by sets of word embeddings.
Likewise, in graph mining, a graph may be viewed as a set of vectors where these vectors correspond to the embeddings of its nodes.
In computer vision, images may be described by local features extracted from different regions of the image.
In such scenarios, one set of feature vectors denotes a single instance of a particular class of interest (an object, document, graph, etc.).
Performing machine learning tasks on such types of objects (\eg set classification, set regression, etc.) is very challenging.
While typical machine learning algorithms are designed for fixed dimensional data instances, the cardinalities of these sets are not fixed, but they are allowed to vary.
Furthermore, the elements of the sets usually do not have an inherent ordering.
Hence, machine learning algorithms have to be invariant to permutations of these elements.

Traditionally, the most common approach to the problem is to define a distance/similarity measure or kernel that finds a correspondence between each pair of sets, and to combine it with an instance-based machine learning algorithm such as $k$-nn or SVM.
This approach has dominated the field, and has achieved state-of-the-art results on many datasets.
However, its main disadvantage is that it is a two-step approach.
Data representation and learning are independent from each other.
Ideally, we would like to have an end-to-end approach.
Besides the above problem, these methods usually suffer from high computational and memory complexity since they compare all sets to each other.

In recent years, neural network architectures have proven extremely successful on a wide variety of tasks, notably in computer vision, in natural language processing, and in graph mining \citep{lecun2015deep}.
One of the main reasons of the success of neural networks is that the representation of data is adapted to the task at hand.
Specifically, recent neural network models are end-to-end trainable, and they generate features that are suitable for the task at hand.
Most models that operate on sets usually update the representations of the elements of the set using some architectures, typically a stack of fully-connected layers.
And then, they apply a permutation invariant function to the updated element representations to obtain a representation for the set.
Although these networks have proven successful in several tasks, they usually employ very simple permutation invariant functions such as the sum or the average of the representations of the elements.
This may potentially limit the expressive power of these architectures.

In this paper, we propose a novel neural network architecture for performing machine learning tasks on sets of vectors, named RepSet.
The network is capable of generating representations for unordered, variable-sized feature sets.
Interestingly, the proposed model produces exactly the same output for all possible permutations of a set of vectors.
To achieve that, it generates a number of hidden sets and it compares the input set with these sets using a network flow algorithm such as bipartite matching.
The outputs of the network flow algorithm form the penultimate layer and are passed on to a fully-connected layer which produces the output.
Since the objective functions of the employed network flow algorithms are differentiable, we can update these hidden sets during training with backpropagation.
Hence, the proposed neural network is end-to-end trainable, while the hidden sets are different for each problem considered.
Deeper models can be obtained by stacking more fully-connected layers one after another.
When the size of the sets is large, solving the flow problems can become prohibitive.
Hence, we also propose a relaxed formulation, ApproxRepSet, which is also permutation invariant and which scales to very large datasets.
We demonstrate the proposed architecture in two classification tasks: text categorization from sets of word embeddings, and graph classification from bags of node embeddings.
The results show that the proposed model produces better or competitive results with state-of-the-art techniques.
Our main contributions are summarized as follows:
\begin{itemize}
    \item We propose RepSet a novel architecture for performing machine learning on sets which, in contrast to traditional approaches, is capable of adapting data representation to the task at hand.
    \item We also propose ApproxRepSet, a simplified architecture which can be interpreted as an approximation of the proposed model, and which can handle very large datasets.
    \item We evaluate the proposed architecture on several benchmark datasets, and achieve state-of-the-art performance.
\end{itemize}

The rest of this paper is organized as follows.
Section \ref{sec:related_work} provides an overview of the related work.
Section \ref{sec:contribution} provides a description of the proposed neural network for sets.
Section \ref{sec:experiments} evaluates the proposed architecture in two different tasks.
Finally, Section \ref{sec:conclusion} concludes.

\section{Related Work}\label{sec:related_work}
The past years witnessed a surge of interest in the area of neural networks for sets.
These networks served mainly as the answer to computer vision problems such as the automated classification of point clouds.
Although conceptually simple, the proposed architectures have achieved state-of-the-art results on many tasks.
PointNet \citep{qi2017pointnet} and DeepSets \citep{zaheer2017deep} transform the vectors of the sets (\ie using several layers) into new representations.
They then apply some permutation-invariant function to the emerging vectors to generate representations for the sets.
PointNet uses max-pooling to aggregate information across vectors, while DeepSets adds up the representations of the vectors.
The representation of the set is then passed on to a standard architecture (\eg fully-connected layers, nonlinearities, etc).
PointNet++ \citep{qi2017pointnet++} and SO-Net \citep{li2018so} apply PointNet hierarchically in order to better capture local structures.
Two other recent works employ neural networks to learn the parameters of the likelihood of each set \citep{rezatofighi2017deepsetnet,rezatofighi2018joint}.
\cite{vinyals2015order} treat unordered sets as ordered sequences and apply RNN models to them.
However, they show that the output of the network is highly dependent on the order of the elements of the set.
More recently, \cite{lee2019set} proposed Set Transformer, a neural network that uses self-attention to model interactions among the elements of the input set.

Besides neural network architectures, several kernels between sets of vectors have been proposed in the past to enable kernel methods (\eg SVMs) to handle unordered sets.
Most of these kernels estimate a probability distribution on each set of vectors, and then derive their similarity using distribution-based comparison measures such as Fisher kernels \citep{jaakkola1999exploiting}, probability product kernels \citep{jebara2004probability,lyu2005kernel} and the classical Bhattacharyya similarity measure \citep{kondor2003kernel}.
Furthermore, there are also kernels that map the vectors of each set to multi-resolution histograms, and then in order to find an approximate correspondence between the two sets of vectors, they compare the histograms with a weighted histogram intersection measure  \citep{grauman2007pyramid,grauman2007approximate}.
Such kernels have been applied to different tasks such as to the problem of graph classification \citep{nikolentzos2017matching}.
Although very effective in several tasks, these kernel-based approaches suffer from high computational complexity.
In most cases, the complexity of computing kernels between sets is quadratic in the number of their elements, while in classification problems, the complexity of optimizing the SVM classifier is quadratic in the number of training samples. 

There are also some very popular permutation-invariant metrics for comparing unordered sets of vectors such as the Earth Mover's distance which corresponds to the solution of an optimization problem that transforms one set to another.
This distance was first introduced by Gaspard Monge in the context of transportation theory, and is often used in computer vision \citep{rubner2000earth} and in natural language processing \citep{kusner2015word}.

\section{Neural Networks for Learning Representations of Sets}\label{sec:contribution}
Conventional machine learning algorithms are designed to operate on fixed-size feature vectors, and they are thus unable to handle sets.
In our setting, each example is represented as a collection $X = \{ \mathbf{v}_1, \mathbf{v}_2, \ldots, \mathbf{v}_n \}$ of $d$-dimensional vectors, $\mathbf{v}_i \in \mathbb{R}^d$.
Note that examples are allowed to vary in the number of elements.
Hence, it is not necessary that all examples comprise of exactly $n$ components.
Since our input is a set $X = \{ \mathbf{v}_1, \mathbf{v}_2, \ldots, \mathbf{v}_n \}$, $\mathbf{v}_i \in \mathbb{R}^d$, our input domain is the power set $\mathcal{X} = 2^{\mathbb{R}^d}$, and we would like to design an architecture whose output would be the same regardless of the ordering of the elements of $X$.
Clearly, to achieve that, a permutation invariant function is necessary to be introduced at some layer of the architecture.
In fact, the model that we propose consists simply of standard fully-connected layers along with a permutation invariant layer.
We next present the proposed permutation invariant layer in detail.

\noindent\textbf{Permutation invariant layer.}
In this paper, we propose a novel permutation invariant layer which capitalizes on well-established concepts from flows and matchings in graphs.
The proposed layer contains $m$ ``hidden sets'' $Y_1, Y_2, \ldots, Y_m$ of $d$-dimensional vectors.
These sets may have different cardinalities and their components are trainable, \ie the elements of a hidden set $Y_i$ correspond to the columns of a trainable matrix $\mathbf{H}^{(i)}$.
Therefore, each column of matrix $\mathbf{H}^{(i)}$ is a vector $\mathbf{u} \in Y_i$.

\tikzset{axis/.style={Black!75!black, -latex, shorten <=0cm, shorten >=-2*\nudge cm}}
\tikzset{line/.style={thick,Green}}

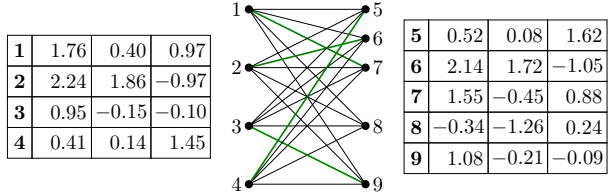
\begin{figure}[t]
    \centering
    \resizebox{\columnwidth}{!}{%
    \begin{tikzpicture}
    
    \tikzstyle{cell} = [rectangle,draw,baseline]
    
    \matrix[matrix,
    matrix of math nodes,
    first column text width/.code={%
      \tikzset{
        column 1/.style={
          nodes={text width=1.2em}
        },
      }
    },
    first column text width=1.3em,
    nodes={cell,minimum height=1.5em, text width=2.9em,
                anchor=center, align=center, %text width=2em,
                inner sep=0pt, outer sep=0pt},
    row sep =-\pgflinewidth,
    column sep = -\pgflinewidth,
    ampersand replacement=\&
  ] at (-1.4,2.5) (am)
  {\bf{1} \& ~~1.76 \& ~~0.40 \& ~~0.97 \\
 \bf{2} \& ~~2.24 \& ~~1.86 \& -0.97 \\
 \bf{3} \& ~~0.95 \& -0.15 \& -0.10 \\
 \bf{4} \& ~~0.41 \& ~~0.14 \& ~~1.45 \\};
    
 \matrix[matrix,
    matrix of math nodes,
    first column text width/.code={%
      \tikzset{
        column 1/.style={
          nodes={text width=1.2em}
        },
      }
    },
    first column text width=1.3em,
    nodes={cell,minimum height=1.5em, text width=2.9em,
                anchor=center, align=center, %text width=2em,
                inner sep=0pt, outer sep=0pt},
    row sep =-\pgflinewidth,
    column sep = -\pgflinewidth,
    ampersand replacement=\&
  ] at (5.4,2.5) (am)
  {\bf{5} \& ~~0.52 \& ~~0.08 \& ~~1.62 \\
   \bf{6} \& ~~2.14 \& ~~1.72 \& -1.05 \\
   \bf{7} \& ~~1.55 \& -0.45 \& ~~0.88 \\
   \bf{8} \& -0.34 \& -1.26 \& ~~0.24 \\
   \bf{9} \& ~~1.08 \& -0.21 \& -0.09 \\};
 
    %\clip (-\nudge ,-\nudge) rectangle (4+\nudge,4+\nudge);
    \draw (1,1) -- (3,1) coordinate (ineq1);
    \draw (1,1) -- (3,2) coordinate (ineq2);
    \draw (1,1) -- (3,3) coordinate (ineq3);
    \draw[line] (1,2) -- (3,1) coordinate (ineq4);
    \draw (1,2) -- (3,2) coordinate (ineq5);
    \draw (1,2) -- (3,3) coordinate (ineq6);
    \draw (1,3) -- (3,1) coordinate (ineq7);
    \draw (1,3) -- (3,2) coordinate (ineq8);
    \draw (1,3) -- (3,3) coordinate (ineq9);
    \draw (1,4) -- (3,1) coordinate (ineq10);
    \draw (1,4) -- (3,2) coordinate (ineq11);
    \draw[line] (1,4) -- (3,3) coordinate (ineq12);
    
    \draw (1,1) -- (3,3.5) coordinate (ineq13);
    \draw (1,2) -- (3,3.5) coordinate (ineq14);
    \draw[line] (1,3) -- (3,3.5) coordinate (ineq15);
    \draw (1,4) -- (3,3.5) coordinate (ineq16);
    
    \draw[line] (1,1) -- (3,4) coordinate (ineq17);
    \draw (1,2) -- (3,4) coordinate (ineq18);
    \draw (1,3) -- (3,4) coordinate (ineq19);
    \draw (1,4) -- (3,4) coordinate (ineq20);

    \foreach \coord/\adj/\name in {
      {(1,4)}/left/1,
      {(1,3)}/left/2,
      {(1,2)}/left/3,
      {(1,1)}/left/4,
      {(3,4)}/right/5,
      {(3,3.5)}/right/6,
      {(3,3)}/right/7,
      {(3,2)}/right/8,
      {(3,1)}/right/9
    } {
      \fill \coord circle (2pt) node[\adj] {\name};
    };
    
    \end{tikzpicture}
    }
    \caption{Example of a bipartite graph generated from $2$ sets of $3$-dimensional vectors, and of its maximum matching $M$. Green color indicates an edge belongs to $M$. The weight of matching $M$ is equal to $16.05$.}
    \label{fig:bipartite_matching}
\end{figure}

\begin{figure*}[t]
    \centering
    \resizebox{1.6\columnwidth}{!}{%
    \includegraphics{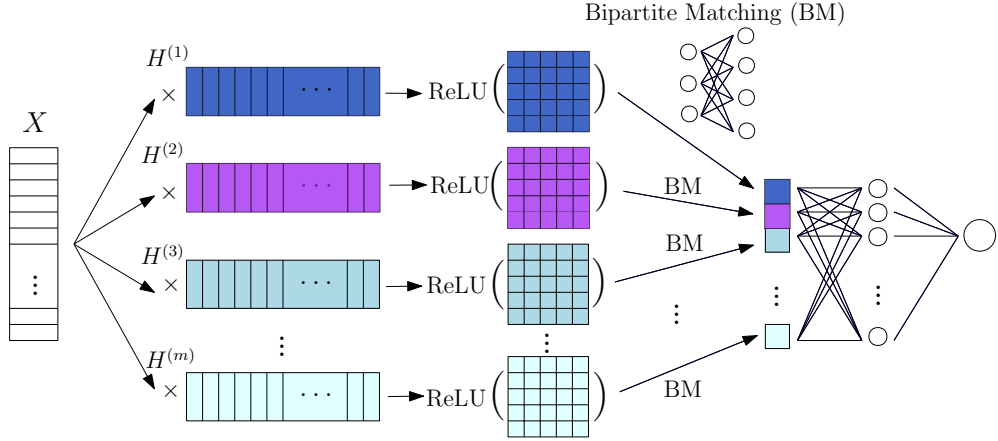}
    }
    \caption{Illustration of the proposed model for learning representations of sets. Each element of the input set is compared with the elements of all ``hidden sets'', and the emerging matrices serve as the input to the bipartite matching algorithm. The objective values of the matching problems correspond to the representation of the input set and are passed on to a standard neural network architecture.}
    \label{fig:architecture}
\end{figure*}

A natural way to measure the similarity between the input set and each one of the hidden sets is by comparing their building blocks, \ie their elements.
To achieve that, we capitalize on network flow algorithms.
Specifically, we use the bipartite matching algorithm to compute a correspondence between the elements of the input set $X$ and the elements of each hidden set $Y_i$.
The bipartite matching problem is one of the most well-studied problems in combinatorial optimization.
The input of the problem is a weighted bipartite graph $G=(V, E)$.
The set of nodes $V$ of a bipartite graph can be decomposed into two disjoint sets $V_1$ and $V_2$, \ie $V=V_1 \cup V_2$.
Every edge $e \in E$ connects a vertex in $V_1$ to one in $V_2$.
A matching $M$ is a subset of edges such that each node in $V$ appears in at most one edge in $M$.
The optimal solution to the problem can be interpreted as the similarity between the two node sets $V_1$ and $V_2$.
A bipartite graph has a natural representation as a rectangular $|V_1| \times |V_2|$ matrix where the $ij^{th}$ component is equal to the weight of the edge between the $i^{th}$ element of $V_1$ and the $j^{th}$ element of $V_2$ if that edge exists, otherwise equal to $0$.
Figure~\ref{fig:bipartite_matching} illustrates a weighted bipartite graph along with the optimal matching $M$.
The weight of each edge (not shown in the Figure) is equal to the inner product of the representations of its endpoints.
Green edges belong to the optimal matching $M$.
In our setting, the input set $X$ corresponds to set $V_1$, the hidden set $Y_i$ corresponds to set $V_2$, and the bipartite graph is complete, \ie every element of $V_1$ is connected to all the elements of $V_2$.
The weight of each edge is the result of a differentiable function $f$ on the representations of the edge's two endpoints.
Formally, given an input set of vectors, $X = \{ \mathbf{v}_1, \mathbf{v}_2, \ldots, \mathbf{v}_{|X|} \}$ and a hidden set $Y = \{ \mathbf{u}_1, \mathbf{u}_2, \ldots, \mathbf{u}_{|Y|} \}$, we can obtain the maximum matching between the elements of the two sets by solving the following linear program:
\begin{equation}
\begin{split}
  \max \displaystyle & \sum_{i=1}^{|X|} \sum_{j=1}^{|Y|} z_{ij} f(\mathbf{v}_i, \mathbf{u}_j) \\
  & \textrm{subject to:} \\
  & \sum_{i=1}^{|X|} z_{ij} \leq 1 \quad \forall j \in \{ 1,\ldots,|Y| \} \\
  & \sum_{j=1}^{|Y|} z_{ij} \leq 1 \quad \forall i \in \{ 1,\ldots,|X| \} \\
  & z_{ij} \geq 0 \quad \forall i \in \{ 1,\ldots,|X| \}, \forall j \in \{ 1,\ldots,|Y| \} \\ 
\end{split}
\label{eq:bipartite_primal}
\end{equation}
where $|X|,|Y|$ are the cardinalities of $X$ and $Y$, $f(\mathbf{v}_i, \mathbf{u}_j)$ is, as mentioned above, a differentiable function, and $z_{ij}=1$ if component $i$ of $X$ is assigned to component $j$ of $Y_i$ and $0$ otherwise.
In our experiments, we have defined $f(\mathbf{v}_i, \mathbf{u}_j)$ as the inner product between the two vectors $\mathbf{v}_i$, and $\mathbf{u}_j$ followed by the ReLU activation function.
Hence, $f(\mathbf{v}_i, \mathbf{u}_j) = \text{ReLU}(\mathbf{v}_i^\top \mathbf{u}_j)$.

Given an input set $X$ and the $m$ hidden sets $Y_1, Y_2, \ldots, Y_m$, we formulate $m$ different bipartite matching problems, and by solving all $m$ of them, we end up with an $m$-dimensional vector $\mathbf{x}$ which corresponds to the hidden representation of set $X$.
This $m$-dimensional vector can be used as features for different machine learning tasks such as set regression or set classification.
For instance, in the case of a set classification problem with $|\mathcal{C}|$ classes, the output is computed as follows:
\begin{equation*}
    \mathbf{p} = \text{softmax}(\mathbf{W} \, \mathbf{x} + \mathbf{b})
\end{equation*}
where $\mathbf{W} \in \mathbb{R}^{|\mathcal{C}| \times m}$ is a matrix of trainable parameters and $\mathbf{b} \in \mathbb{R}^{|\mathcal{C}|}$ is the bias term.
Given a training set consisting of sets $X_1, X_2, \ldots, X_N$, we use the negative log likelihood of the correct labels as training loss:
\begin{equation*}
    L = -\sum_{i=1}^N \sum_{j=1}^{|\mathcal{C}|} \mathbf{y}_j^i \log \mathbf{p}_j^i
\label{eq:loss_function}    
\end{equation*}
where $\mathbf{y}_j^i$ is equal to $1$ if $X_i$ belongs to the $j^{th}$ class, and $0$ otherwise.
Note that we can create a deeper architecture by adding more fully-connected layers.
An overview of the proposed architecture is illustrated in Figure~\ref{fig:architecture}.

As mentioned above, a constraint that the neural network is necessary to satisfy is to be invariant under any permutation of the elements of the input set.
Our next theoretical result shows that the proposed model generates the same output for all $n!$ permutations of the elements of an input set $X$.
\begin{theorem}
    Let $X$ be a set having elements from a countable or uncountable universe.
    The proposed architecture is invariant to the permutation of elements in $X$.
\end{theorem}
%\begin{proof}
%The proof is left to the supplementary material.
%\end{proof}

\textbf{Computing the derivative with respect to the hidden sets.}
The objective function of the proposed architecture is differentiable, and we can find a minimum by using classical stochastic optimization techniques that have proven very successful for training deep neural networks.
We next use the chain rule and show how the gradients are computed.
Note that here we assume a multiclass classification setting.
The gradients for different settings (\eg regression) are computed in a similar fashion.

Let $\mathbf{r} = \mathbf{W} \mathbf{x} + \mathbf{b}$.
We use the chain rule in order to compute the derivative with respect to the weight matrix $\mathbf{H}^{(k)}$ of hidden set $Y_k$:
\begin{equation}
    \frac{\partial L}{\partial \mathbf{H}^{(k)}} = \frac{\partial L}{\partial \mathbf{r}}\frac{\partial \mathbf{r}}{\partial \mathbf{x}_k} \frac{\partial \mathbf{x}_k}{\partial \mathbf{H}^{(k)}}
    \label{eq:der}
\end{equation}
where $\mathbf{x}_k$ is the $k^{th}$ component of vector $\mathbf{x}$.
The gradient of the loss function with respect to $\mathbf{r}$ is:
\begin{equation}
    \frac{\partial L}{\partial \mathbf{r}} = \mathbf{p} - \mathbf{y}
    \label{eq:der1}
\end{equation}
We also have that:
\begin{equation}
    \frac{\partial \mathbf{r}}{\partial \mathbf{x}_k} = \mathbf{W}_k
    \label{eq:der2}
\end{equation}
where $\mathbf{W}_k$ denotes the $k^{th}$ column of matrix $\mathbf{W}$.
The $k^{th}$ component of vector $\mathbf{x}$ is equal to:
\begin{equation*}
    \mathbf{x}_k = \sum_{i=1}^{|X|} \sum_{j=1}^{|Y_k|}\mathbf{D}^{(k)}_{ij}f(\mathbf{v}_i,\mathbf{u}^{(k)}_j) = \Tr(\mathbf{D}^{(k)^{\top}} \, \mathbf{F}^{(k)})
\end{equation*}
where $\mathbf{D}^{(k)}$ is the matrix that contains the optimal values of the variables of problem (\ref{eq:bipartite_primal}) for $Y_k$, \ie $\mathbf{D}^{(k)}_{ij} = z_{ij}^{(k)}$, $\mathbf{u}^{(k)}_j$ is the $j^{th}$ column of matrix $\mathbf{H}^{(k)}$, and $\mathbf{F}^{(k)} \in \mathbb{R}^{|X| \times |Y_k|}$ is a matrix such that $\mathbf{F}^{(k)}_{ij} = f(\textbf{v}_i,\textbf{u}_j)$.

Let $\mathbf{V}$ be a matrix whose rows correspond to the elements of set $X$.
Then, it holds that $\mathbf{F}^{(k)}=\text{ReLU}(\mathbf{V}\mathbf{H}^{(k)})$.
This yields:
\begin{equation}
    \frac{\partial \mathbf{x}_k}{\partial \mathbf{H}^{(k)}} = \frac{\partial }{\partial \mathbf{H}^{(k)}} \Tr \big(\mathbf{D}^{(k)^{\top}} \, \text{ReLU}(\mathbf{V}\mathbf{H}^{(k)}) \big) = \mathbf{V}^\top \mathbf{D}^{(k)}
    \label{eq:der3}
\end{equation}
From Equations~(\ref{eq:der}), (\ref{eq:der1}), (\ref{eq:der2}) and (\ref{eq:der3}), we finally have: 
\begin{equation*}
    \frac{\partial L}{\partial \mathbf{H}^{(k)}} = \frac{\partial L}{\partial \mathbf{r}}\frac{\partial \mathbf{r}}{\partial \mathbf{x}_k} \frac{\partial \mathbf{x}_k}{\partial \mathbf{H}^{(k)}} = (\mathbf{p} - \mathbf{y})^\top \, \mathbf{W}_k \, (\mathbf{V}^\top \, \mathbf{D}^{(k)})
\end{equation*}

\noindent\textbf{Relaxed problem.}
The major weakness of the above architecture is its computational complexity.
Computing a maximum cardinality matching in a weighted bipartite graph with $n$ vertices and $m$ edges takes time $\mathcal{O}(mn + n^2 \log n)$, using the classical Hungarian algorithm.
This prohibits the proposed model from being applied to very large datasets.
To account for that, we next present ApproxRepSet, an approximation of the bipartite matching problem which involves operations that can be performed on a GPU, allowing thus efficient implementations.
More specifically, given an input set of vectors, $X = \{ \mathbf{v}_1, \mathbf{v}_2, \ldots, \mathbf{v}_{|X|} \}$ and a hidden set $Y = \{ \mathbf{u}_1, \mathbf{u}_2, \ldots, \mathbf{u}_{|Y|} \}$, first we identify which of the two sets has the highest cardinality.
If $|X| \geq |Y|$, we solve the following linear program:
\begin{equation}
\begin{split}
  \max \displaystyle & \sum_{i=1}^{|X|} \sum_{j=1}^{|Y|} z_{ij} f(\mathbf{v}_i, \mathbf{u}_j) \\
  & \textrm{subject to:} \\
  & \sum_{i=1}^{|X|} z_{ij} \leq 1 \quad \forall j \in \{ 1,\ldots,|Y| \} \\
  & z_{ij} \geq 0 \quad \forall i \in \{ 1,\ldots,|X| \}, \forall j \in \{ 1,\ldots,|Y| \} \\ 
\end{split}
\label{eq:approx_primal}
\end{equation}

Conversely, if $|X| < |Y|$, then we replace the first constraint with the following: $\sum_{j=1}^{|Y|} z_{ij} \leq 1 \quad \forall i \in \{ 1,\ldots,|X| \} $.
This problem is clearly a relaxed formulation of the problem defined in Equation~(\ref{eq:bipartite_primal}) where a constraint has been removed.
\begin{proposition}
    The optimization problem defined in Equation~\ref{eq:approx_primal} is an upper bound to the bipartite matching problem defined in Equation~\ref{eq:bipartite_primal}.
\end{proposition}

\section{Experimental Evaluation}\label{sec:experiments}
We next evaluate the proposed model in text categorization and graph classification, and compare it against strong baselines.
Further experimental results are provided in the supplementary material.

\subsection{Text Categorization}
We first evaluate RepSet and ApproxRepSet in the task of text categorization.
Given a document, the input to the model is the set of embeddings of its terms.

\noindent\textbf{Baselines.}
We compare the proposed architecture against Word Mover's Distance (WMD) \citep{kusner2015word} and its supervised version (S-WMD) \citep{huang2016supervised}, against DeepSets \citep{zaheer2017deep} and three variants of it, where we replaced the sum operator with mean (NN-mean) and max (NN-max) operators, and with an attention mechanism (NN-attention), and against Set-Transformer \citep{lee2019set}.

\begin{table*}[t] 
\centering
\caption{Classification test error of the proposed architecture and the baselines on $8$ text categorization datasets.}
\resizebox{2\columnwidth}{!}{
	\begin{tabular}{l c c c c c c c c}
    \toprule
      & BBCSPORT & TWITTER & RECIPE & OHSUMED & CLASSIC & REUTERS & AMAZON & 20NG \\
    \midrule
    WMD & 4.60 $\pm$ 0.70 & 28.70 $\pm$ 0.60 & 42.60 $\pm$ 0.30 & 44.50 & {\bf 2.88} $\pm$ 0.10 & 3.50 & 7.40 $\pm$ 0.30 & 26.80 \\
    S-WMD & 2.10 $\pm$ 0.50 & 27.50 $\pm$ 0.50 & 39.20 $\pm$ 0.30 & 34.30 & 3.20 $\pm$ 0.20 & 3.20 & 5.80 $\pm$ 0.10 & 26.80 \\
    \cmidrule{1-9}
    DeepSets & 25.45 $\pm$ 20.1 & 29.66 $\pm$ 1.62 & 70.25 $\pm$ 0.00 & 71.53 & 5.95 $\pm$ 1.50 & 10.00 & 8.58 $\pm$ 0.67 & 38.88 \\
    NN-mean & 10.09 $\pm$ 2.62 & 31.56 $\pm$ 1.53 & 64.30 $\pm$ 7.30 & 45.37 & 5.35 $\pm$ 0.75 & 11.37 & 13.66 $\pm$ 3.16 & 38.40\\
    NN-max & 2.18 $\pm$ 1.75 & 30.27 $\pm$ 1.26 & 43.47 $\pm$ 1.05 & 35.88 & 4.21 $\pm$ 0.11 & 4.33 & 7.55 $\pm$ 0.63 & 32.15 \\
    NN-attention & 4.72 $\pm$ 0.97 & 29.09 $\pm$ 0.62 & 43.18 $\pm$ 1.22 & {\bf 31.36} & 4.42 $\pm$ 0.73 & 3.97 & 6.92 $\pm$ 0.51 & 28.73\\
    \cmidrule{1-9}
    Set-Transformer & 4.18 $\pm$ 1.23 & 27.79 $\pm$ 0.47 & 42.54 $\pm$ 1.35 & 35.68 & 5.23 $\pm$ 0.52 & 4.52 & 7.18 $\pm$ 0.44 & 30.01 \\
    \cmidrule{1-9}
    RepSet & {\bf 2.00} $\pm$ 0.89 & {\bf 25.42} $\pm$ 1.10 & {\bf 38.57} $\pm$ 0.83 & 33.88 & 3.38 $\pm$ 0.50 & 3.15 & {\bf 5.29} $\pm$ 0.28 & {\bf 22.98} \\
    ApproxRepSet & 4.27 $\pm$ 1.73 & 27.40 $\pm$ 1.95 & 40.94 $\pm$ 0.40 & 35.94 & 3.76 $\pm$ 0.45 & \bf 2.83 & 5.69 $\pm$ 0.40 & 23.82 \\
    %ApproxRepSet & 8.45 $\pm$ 2.27 & 25.51 $\pm$ 1.03 & 43.26 $\pm$ 0.13 & 42.90 & 4.08 $\pm$ 0.76 & 3.83 & 6.27 & 29.18\\
    \bottomrule
	\end{tabular}
  }
\label{tab:text_categorization}
\end{table*}

\begin{table*}[t] 
  \centering
  \caption{Terms of the employed pre-trained model that are most similar to the elements and centroids of $5$ hidden sets.}
  \resizebox{1.4\columnwidth}{!}{
  \begin{tabular}{c c c}
  	\toprule
      Hidden & Terms similar to & Terms similar to \\
      set & elements of hidden sets & centroids of hidden sets \\
      \midrule
      1 & chelsea, football, striker, club, champions & footballing \\
       \cmidrule{1-3}
      2 & qualify, madrid, arsenal, striker, united, france & ARSENAL\_Wenger\\
       \cmidrule{1-3}
      3 & olympic, athlete, olympics, sport, pentathlon & Olympic\_Medalist \\
      \cmidrule{1-3}
      4 & penalty, cup, rugby, coach, goal & rugby \\
       \cmidrule{1-3}
      5 & match, playing, batsman, batting, striker & batsman\\
           \bottomrule
  \end{tabular}
  }
  \label{tab:latent_sets}
\end{table*}

\noindent\textbf{Data and setup.}
We evaluate all approaches on $8$ document classification datasets: ($1$) BBCSPORT, ($2$) TWITTER, ($3$) RECIPE, ($4$) OHSUMED, ($5$) CLASSIC, ($6$) REUTERS, ($7$) AMAZON, ($8$) 20NG.
More details about the datasets are given in the supplementary material.

For the proposed models, we use a two-layer architecture: a permutation invariant layer followed by a fully-connected layer.
Code is available at: \url{https://github.com/giannisnik/repset}.
We choose the number of hidden sets from $\{ 20,30,50,100 \}$ and their cardinality from $\{ 10,20,50 \}$ based on validation experiments.
For DeepSets and its three variants, we use $3$ fully-connected layers of sizes $\{300,100,30\}$ with tanh activations, followed by the aggregation operator, and then by $2$ fully-connected layers of sizes $\{30, 10\}$ with tanh activations.
For Set-Transformer, we use a stack of set attention blocks in encoder and a pooling by multihead attention module followed by a stack of set attention blocks in decoder.
The dimensionality of all hidden layers is set to $64$ and the number of attention heads to $4$.
For WMD and S-WMD, we provide the results reported in the original papers.

\noindent\textbf{Results.}
Table~\ref{tab:text_categorization} shows the average classification error of the proposed models and those of the baselines.
On all datasets except two (OHSUMED, CLASSIC), the proposed model outperforms the baselines.
In some cases, the gains in accuracy over the best performing competitors are considerable.
For instance, on the 20NG, TWITTER, and RECIPE datasets, RepSet achieves respective absolute improvements of $3.82\%$, $2.08\%$ and $0.63\%$ in accuracy over the best competitor, the S-WMD method.
Furthermore, the proposed model outperforms DeepSets and its variants, and Set-Transformer on all datasets, and on most of them by wide margins.
Overall, it is clear from Table~\ref{tab:text_categorization} that RepSet is superior to the other methods in text categorization.
Regarding the two variants of the proposed architecture, the model that solves exactly the bipartite matching problem (RepSet) outperforms the model that approximates it (ApproxRepSet) on all datasets except from REUTERS.
However, on most datasets the difference in performance is not large.
Hence, although less powerful, ApproxRepSet is still capable of learning expressive representations of sets.

We should mention at this point that besides effective, the proposed model is also highly interpretable.
For instance, in the text categorization setting, the elements of each hidden set can be regarded as the terms of hidden documents which are likely to be related to the topics of the different classes.
To experimentally verify that, we trained a model with $50$ hidden sets on the BBCSPORT dataset.
Each hidden set consisted of $20$ vectors.
For $5$ hidden sets, we found the terms that are closer to their elements and we also computed their centroids.
Table~\ref{tab:latent_sets} shows the terms of the pre-trained model that were found to be most similar (using cosine similarity), with respect to the vectors and centroids.
Clearly, the centroids of these $5$ hidden sets are close to terms that are related to sports.
Interestingly, some of these terms correspond to cricket positions, while others to names of famous soccer teams.

\begin{table*}[t]
    \centering
    \caption{Classification accuracy ($\pm$ standard deviation) of the proposed architecture and the baselines on the $5$ graph classification datasets.}
    \resizebox{1.6\columnwidth}{!}{
      \begin{tabular}{l c c c c c}
      \toprule
      & \multirow{2}{*}{\textsc{MUTAG}} &  \multirow{2}{*}{\textsc{PROTEINS}} & \textsc{IMDB} & \textsc{IMDB} & \textsc{REDDIT}  \\
      & & & BINARY & MULTI & BINARY \\ 
      \midrule
      PSCN $k=10$ & 88.95 ($\pm$ 4.37) & 75.00 ($\pm$ 2.51) & 71.00 ($\pm$ 2.29) &  45.23 ($\pm$ 2.84) & 86.30 ($\pm$ 1.58) \\
      Deep GR & 82.66 ($\pm$ 1.45) & 71.68 ($\pm$ 0.50) & 66.96 ($\pm$ 0.56) & 44.55 ($\pm$ 0.52) &  78.04 ($\pm$ 0.39) \\
      EMD & 86.11 ($\pm$ 0.84) & - & - & - & -  \\
      DGCNN &  85.80 ($\pm$ 1.70) & 75.50 ($\pm$ 0.90) & 70.03 ($\pm$ 0.86) & 47.83 ($\pm$ 0.85) & -  \\
      SAEN & 84.99 ($\pm$ 1.82) & 75.31 ($\pm$ 0.70) & 71.59 ($\pm$ 1.20) & 48.53 ($\pm$ 0.76) & 87.22 ($\pm$ 0.80) \\
      RetGK & {\bf90.30} ($\pm$ 1.10) & 76.20 ($\pm$ 0.50) & 72.30 ($\pm$ 0.60) & 48.70 ($\pm$ 0.60) & {\bf 92.60} ($\pm$ 0.30) \\
      DiffPool & - & {\bf 76.25} & - & - & - \\
      \cmidrule{1-6}
      DeepSets & 86.26 ($\pm$ 1.09) & 60.82 ($\pm$ 0.79) & 69.84 ($\pm$ 0.64) & 47.62 ($\pm$ 1.18) & 52.01 ($\pm$ 1.47) \\
      NN-mean & 87.55 ($\pm$ 0.98) & 73.00 ($\pm$ 1.21) & 71.48 ($\pm$ 0.48) & 49.92 ($\pm$ 0.82) & 84.57 ($\pm$ 0.84) \\
      NN-max & 85.84 ($\pm$ 0.99) & 71.05 ($\pm$ 0.54) & 69.56 ($\pm$ 0.91) & 48.28 ($\pm$ 0.43) & 80.98 ($\pm$ 0.79)\\
      NN-attention & 85.92 ($\pm$ 1.16) & 74.48 ($\pm$ 0.22) & {\bf 72.40} ($\pm$ 0.45) & 49.56 ($\pm$ 0.47) & 88.74 ($\pm$ 0.53)\\
      \cmidrule{1-6}
    Set-Transformer & 87.71 ($\pm$ 1.14) & 59.62 ($\pm$ 1.42) & 71.21 ($\pm$ 1.28) & {\bf 50.25} ($\pm$ 0.74) & 83.79 ($\pm$) 0.83\\
     \cmidrule{1-6}
      RepSet & 88.63 ($\pm$ 0.86) & 73.04 ($\pm$ 0.42) & {\bf 72.40} ($\pm$ 0.73) & 49.93 ($\pm$ 0.60) & 87.45 ($\pm$ 0.86)\\
      %\cmidrule{1-6}
      ApproxRepSet & 86.33 ($\pm$ 1.48) & 70.74 ($\pm$ 0.85) & 71.46 ($\pm$ 0.91) & 48.92 ($\pm$ 0.28) & 80.30 ($\pm$ 0.56) \\
       \bottomrule
      \end{tabular}
    }
    \label{tab:graph_classification}
\end{table*}

\subsection{Graph Classification}
We also evaluate the proposed architecture in the graph classification task.
We represent each graph as a set of vectors (\ie the embeddings of its nodes), and pass them on to the proposed models.

\noindent\textbf{Baselines.}
We compare the proposed models against several recent state-of-the-art approaches: ($1$) PSCN, a model that extracts neighborhood subgraphs of specific size, defines an ordering on the nodes of these subgraphs and feeds the emerging adjacency matrices into a convolutional neural network \citep{niepert2016learning}, ($2$) Deep GR, an approach that improves the graphlet kernel by using the Skip-gram model \citep{yanardag2015deep}, ($3$) EMD, a method that represents each graph as a set of vectors and computes the distance between each pair of graphs using the Earth Mover's Distance algorithm \citep{nikolentzos2017matching}, ($4$) DGCNN, a model that applies a message passing architecture followed by a pooling operator based on sorting to create a fixed-sized graph representation which is then passed on to a convolutional architecture \citep{zhang2018end}, ($5$) SAEN, an algorithm that decomposes graphs in a hierarchical fashion, then uses shift, aggregate and extract operations, and finally applies a standard neural network to the emerging representations \citep{orsini2018shift}, ($6$) RetGK, a graph kernel that capitalizes on the isomorphism-invariance property of the return probabilities of random walks \citep{zhang2018retgk}, and ($7$) DiffPool, a message passing architecture which applies at each layer a differentiable graph pooling module that clusters the nodes of the previous layer \citep{ying2018hierarchical}.
We also compare the proposed models against DeepSets \citep{zaheer2017deep} and the three variants of it that were presented above (NN-mean, NN-max, and NN-attention), and against Set-Transformer \citep{lee2019set}.

\noindent\textbf{Data and setup.}
We evaluate the competing methods on standard graph classification datasets derived from bioinformatics (MUTAG, PROTEINS) and social networks (IMDB-BINARY, IMDB-MULTI, REDDIT-BINARY) \citep{KKMMN2016}.
Note that the social network graphs are unlabeled, while the bioinformatics graphs contain node labels.
However, in our experiments, we did not take these node labels into account.
More details about the datasets are given in the supplementary material.
Since the datasets do not come with standard training/test splits, we performed $10$-fold cross validation procedure where we randomly sampled $10\%$ of each training fold to serve as a validation set.
Furthermore, since some datasets are small, we repeated the whole experiment $10$ times.

We generated embeddings for the nodes as follows: we create a single graph from each dataset by computing the disjoint union of all the graphs contained in it.
In other words, for each dataset, we generate a disconnected graph whose components correspond to the graphs contained in the dataset.
We then employ struc2vec, an algorithm that learns structural node representations \citep{ribeiro2017struc2vec}.
Nodes with structurally similar neighborhoods are close to each other in the embeddings space.
We set the dimensionality of the learned embeddings to $20$.
For RepSet, ApproxRepSet, DeepSets and its variants, and Set-Transformer, we use the same configuration as in the case of text categorization.
Note that all the above models are trained on the same input data (\ie sets of node embeddings).
For the remaining baseline methods, we provide the results reported in the original papers.

\noindent\textbf{Results.}
Table~\ref{tab:graph_classification} shows the average classification accuracy of the proposed models and those of the baselines.
We observe that the proposed models are on par with the state-of-the-art algorithms.
Specifically, RepSet obtains the highest average performance among all methods on the IMDB-BINARY dataset, while it is the second best method on IMDB-MULTI, and the third best method on REDDIT-BINARY and on MUTAG.
DeepSets and its variants achieve comparable accuracies to the proposed models.
Specifically, the best-performing variant, NN-attention, outperforms the proposed models on PROTEINS and REDDIT-BINARY.
Set-Transformer, on the other hand, outperforms the proposed models only on  IMDB-MULTI.
Interestingly, even though the embeddings which the proposed model utilizes do not incorporate information about the node labels, on the MUTAG dataset (whose graphs contain node labels), it remains competitive with the baselines which take these node labels into account.
On the other hand, on the second dataset that contains node labels (PROTEINS), our proposed models, RepSet and ApproxRepSet, are outperformed by all the baselines except DeepSets, NN-max and Set-Transformer.
Last, our approximate model, ApproxRepSet, achieves comparable to state-of-the-art results in almost all datasets, while being considerably faster than RepSet.

\begin{figure*}[t]
    \centering
    \begin{minipage}{\columnwidth}
    \centering
    \resizebox{0.89\columnwidth}{!}{%
    \includegraphics{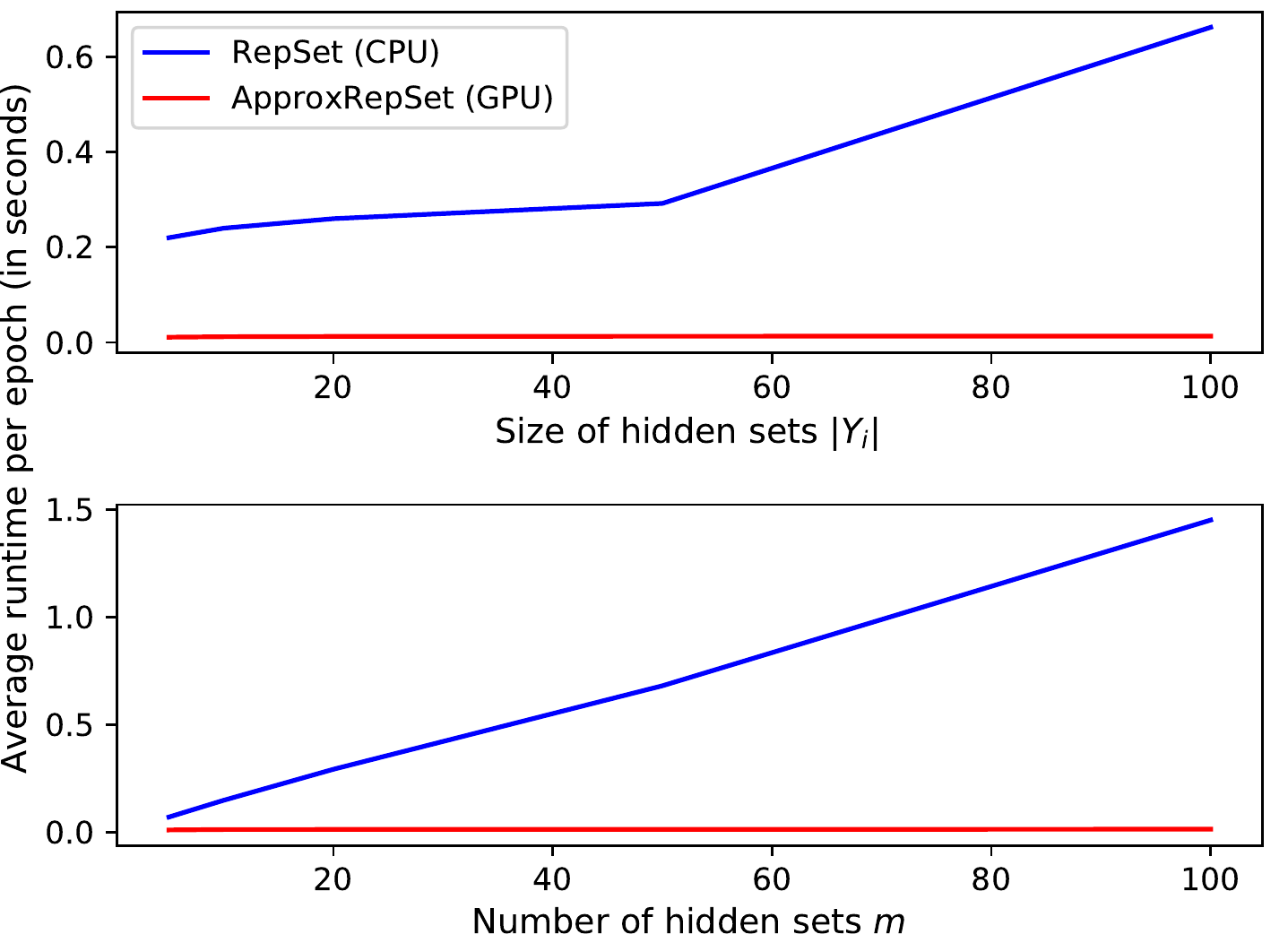}
    }
    \end{minipage}%
    \begin{minipage}{\columnwidth}
        \centering
        \resizebox{0.89\columnwidth}{!}{%
        \includegraphics{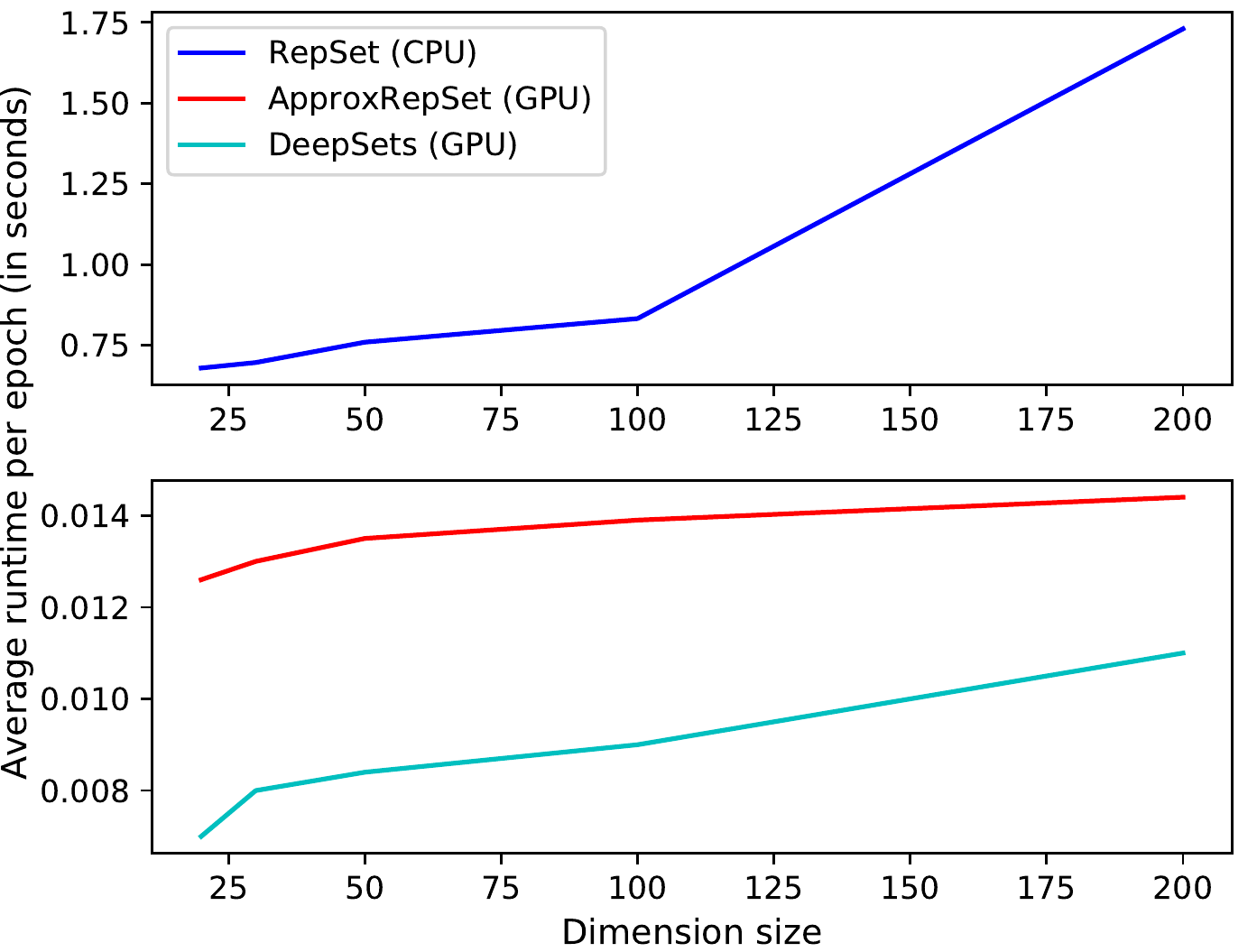}
        }
    \end{minipage}
    \caption{Runtimes with respect to the number of hidden sets $m$, the size of the hidden sets $|Y_i|$ (left) and embeddings with different dimensions (right).}
    \label{fig:runtime_analysis}
    \end{figure*}
    \begin{figure*}[t]
    \centering
        \begin{minipage}{\columnwidth}
        \centering
        \resizebox{0.89\columnwidth}{!}{%
        \includegraphics{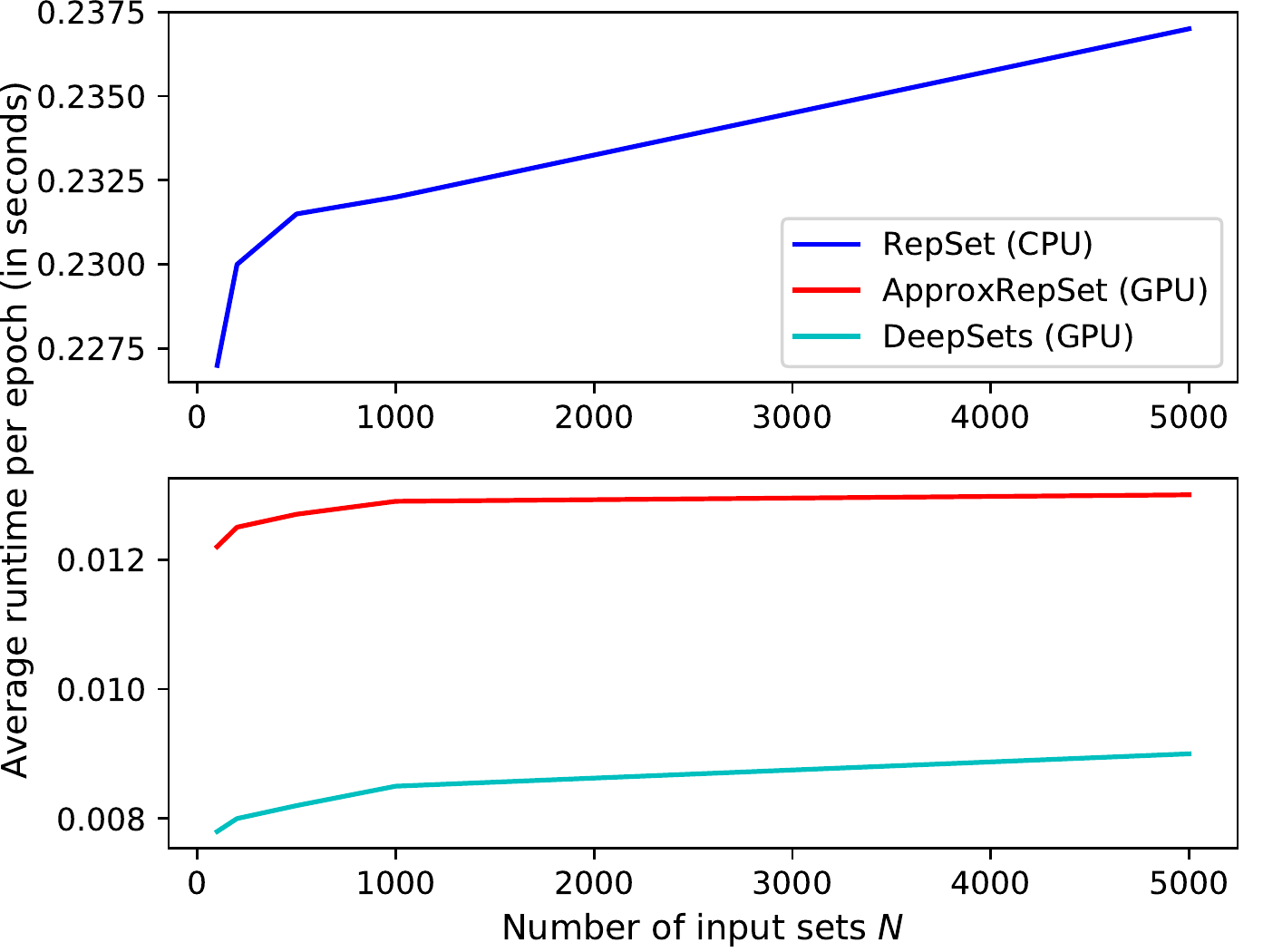}
        }
    \end{minipage}%
    \begin{minipage}{\columnwidth}
         \centering
        \resizebox{0.89\columnwidth}{!}{%
        \includegraphics{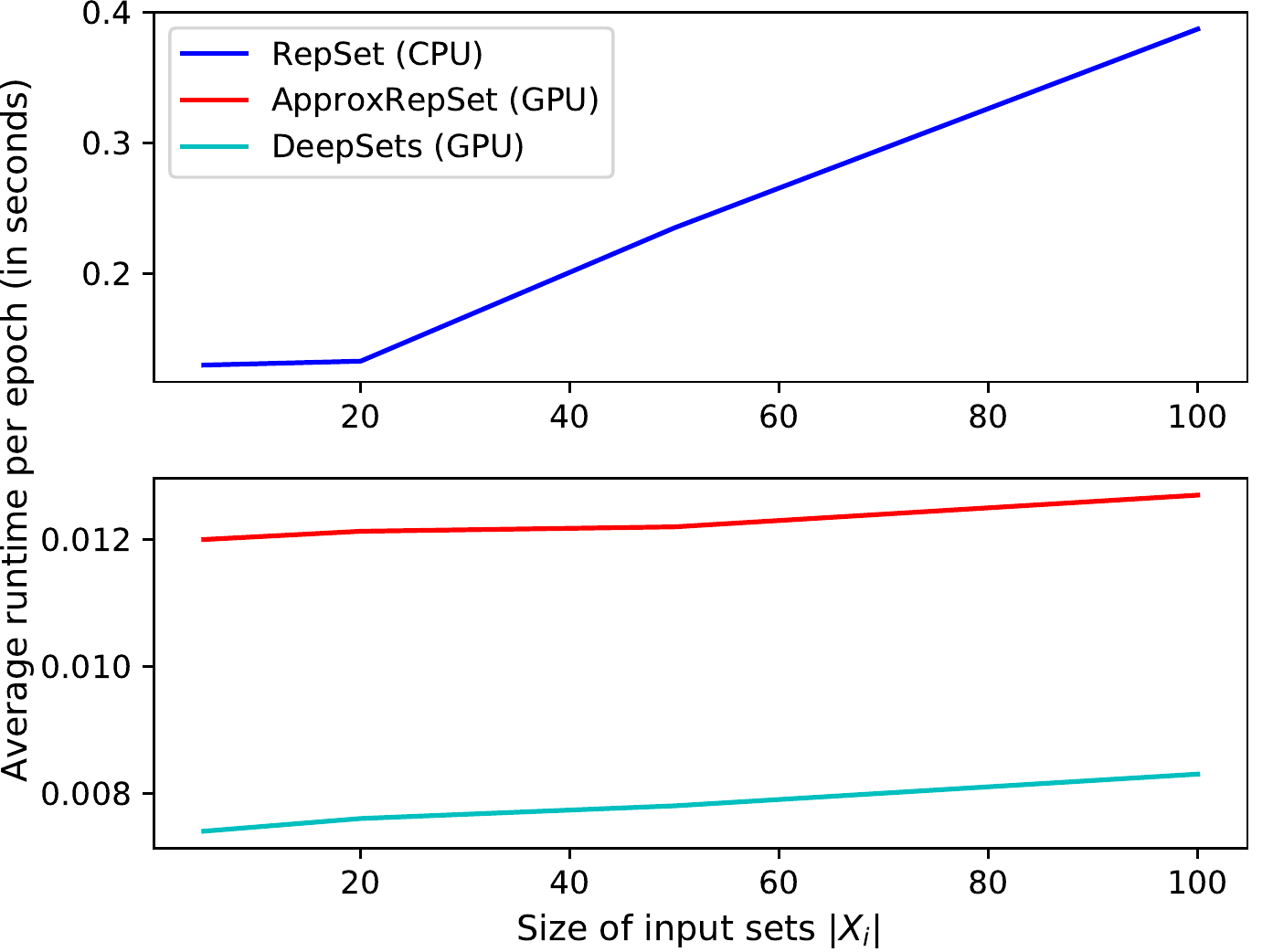}
        }
    \end{minipage}
    \caption{Runtimes with respect to the number of input sets $N$ (left) and the size of the input sets $|X_i|$ (right).}
    \label{fig:runtime_analysis2}
\end{figure*}

\subsection{Runtime Analysis}
To evaluate the runtime performance and scalability of the proposed models, we created a series of synthetic datasets and measured how the average running time per epoch varies with respect to the parameters of the model.
Figure~\ref{fig:runtime_analysis} illustrates the running time of RepSet and ApproxRepSet as a function of the size of the hidden sets $|Y_i|, i=1,\ldots,m$ (top left) and as a function of the number of hidden sets $m$ (bottom left). 
Note that for RepSet, training was performed on an Intel Xeon E$5-1607$ CPU ($4$ threads), while for ApproxRepSet, it was performed on an NVIDIA Titan Xp GPU.
As expected, we observe that RepSet is more computationally expensive than ApproxRepSet, while its running time increases significantly as the size and the number of hidden sets increase.
On the other hand, the values of these parameters do not have a large impact on the running time of ApproxRepSet.
We also evaluate (Figure~\ref{fig:runtime_analysis} (right)) how the proposed models scale as the dimensionality of the vectors contained in the sets increases and compare them against the DeepSets model.
DeepSets is the fastest model, followed by ApproxRepSet.
The running times of both these models are much smaller than that of RepSet, while they also grow very slowly as the dimensionality of vectors increases.
Conversely, the running time of RepSet increases notably especially for dimensionalities larger than $100$.
We also examine in Figure~\ref{fig:runtime_analysis2} how the running time of the three models varies with respect to the number of input samples $N$ (left) and to their cardinality $|X_i|, i=1,\ldots,N$ (right).
Surprisingly, we find that the running time of RepSet grows slowly as the number of samples increases.
On the other hand, it increases significantly as the cardinality of these samples increases.
The running times of DeepSets and ApproxRepSet are again much lower than that of RepSet, and grow only slightly as the number of samples and their cardinality increase.

\section{Conclusion}\label{sec:conclusion}
In this paper, we proposed RepSet, a neural network for learning set representations.
RepSet computes the correspondences between the input sets and some hidden sets by solving a series of matching/network flow problems.
We also introduced a relaxed version which involves fast matrix operations and scales to large datasets.
Experiments in two tasks show that our architecture is competitive with the state-of-the-art.

\section*{Acknowledgments}
The authors would like to thank the AISTATS reviewers for their insightful comments.
GN was supported by the project ``ESIGMA'' (ANR-17-CE40-0028).

\bibliography{biblio}

\clearpage
\appendix

\section*{Supplementary Material}

\section{Overview}
This document is supplementary material for the paper ``Rep the Set: Neural Networks for Learning Set Representations''.
It is organized as follows.
We will prove in Section~\ref{sec:proof_theorem} the Theorem $1$.
In Section~\ref{sec:proof_proposition}, we will present the proof of Proposition 1.
In Section~\ref{sec:datasets}, we will give details about the datasets we used in our experiments.
Finally, in Section~\ref{sec:further_experiments}, we perform a sensitivity analysis, and we present the features that the model learns on a simple synthetic dataset.

\section{Proof of Theorem 1}\label{sec:proof_theorem}
For the reader's convenience we will restate Theorem 1 from the article.
\begin{theorem}
    Let $X$ be a set having elements from a countable or uncountable universe.
    The proposed architecture is invariant to the permutation of elements in $X$.
\end{theorem}
\begin{proof}
Let $\Pi_{n}$ be the set of all permutations of the integers from $1$ to $n$.
Let $\pi\in\Pi_{|X|}$ be an arbitrary permutation.
We will apply $\pi$ to the input set $X$.
The bipartite matching problem then becomes:
\begin{equation}
\begin{split}
  \max \displaystyle & \sum_{i=1}^{|X|} \sum_{j=1}^{|Y|} z_{ij} f(\mathbf{v}_{\pi(i)}, \mathbf{u}_j) \\
  & \textrm{subject to:} \\
  & \sum_{i=1}^{|X|} z_{ij} \leq 1 \quad \forall j \in \{ 1,\ldots,|Y| \} \\
  & \sum_{j=1}^{|Y|} z_{ij} \leq 1 \quad \forall i \in \{ 1,\ldots,|X| \} \\
  & z_{ij} \geq 0 \quad \forall i \in \{ 1,\ldots,|X| \}, \forall j \in \{ 1,\ldots,|Y| \} \\ 
\end{split}
\label{eq:premutation}
\end{equation}
The constraints of the optimization problem remain intact since summing the elements of a set is a permutation invariant function.
Moreover, it holds that: 
\begin{equation}
    \sum_{i=1}^{|X|} \sum_{j=1}^{|Y|} z_{ij} f(\mathbf{v}_{\pi(i)}, \mathbf{u}_j)=\sum_{i=1}^{|X|} \sum_{j=1}^{|Y|} z_{\pi^{-1}(i)j} f(\mathbf{v}_i, \mathbf{u}_j)
\end{equation}
The sets of variables that lead to the optimal solutions of problems (\ref{eq:bipartite_primal}) in the main paper and (\ref{eq:premutation}) above are identical, \ie $z^*_{\pi^{-1}(i)j}= z^*_{ij}$, $\forall i \in \{ 1,\ldots,|X| \}, \forall j \in \{ 1,\ldots,|Y| \}$.
Hence, the optimal value of the bipartite matching problem is the same for all $n!$ permutations of the input, and therefore, the proposed model maps all permutations of the input into the same representation.
\end{proof}

\section{Proof of Proposition 1}\label{sec:proof_proposition}
For the reader's convenience we will restate Proposition 1 from the article.
\begin{proposition}
    The optimization problem defined in Equation~\ref{eq:approx_primal} is an upper bound to the bipartite matching problem defined in Equation~\ref{eq:bipartite_primal}.
\end{proposition}
\begin{proof}
We assume without loss of generality that $|X| \geq |Y|$.
Let $\mathbf{D}^*$ be the optimal solution to the relaxed problem, \ie $\mathbf{D}_{ij}^* = z_{ij}$.
Therefore, it holds that:
\[ \mathbf{D}_{ij}^*=
  \begin{cases}
    1  & \quad \text{if } i=\argmax_k f(\textbf{v}_k,\textbf{u}_j) \wedge \max_k f(\textbf{v}_k,\textbf{u}_j) > 0\\
    0  & \quad \text{otherwise}
  \end{cases}
\]
The relaxed problem is allowed to match multiple elements of $Y$ with the same element of $X$.
Specifically, the optimal solution of the relaxed problem matches an element of $Y$ with an element of $X$ if their inner product is positive and is the highest among the inner products between that element of $Y$ and all the elements of $X$.
Then, for every $j$, let $i^*=\argmax_k f(\textbf{v}_k,\textbf{u}_j)$.
For any feasible solution $\mathbf{D}$ of the exact problem, and for any $j$, we have:
\begin{equation}
    \begin{split}
        \sum_{i=1}^{|X|} \mathbf{D}_{ij} f(\mathbf{v}_i, \mathbf{u}_j) &\leq \sum_{i=1}^{|X|} \mathbf{D}_{ij} f(\mathbf{v}_{i^*}, \mathbf{u}_j) \\
        &= f(\mathbf{v}_{i^*}, \mathbf{u}_j)\sum_{i=1}^{|X|} \mathbf{D}_{ij} \\ 
        &\leq f(\mathbf{v}_{i^*}, \mathbf{u}_j) \\
        &= \sum_{i=1}^{|X|} \mathbf{D}^*_{ij} f(\mathbf{v}_i, \mathbf{u}_j)
    \end{split}
\end{equation}
Therefore, the objective value of the relaxed problem (obtained by $\mathbf{D}^*$) gives an upper to the exact problem.
\end{proof}

\section{Datasets}\label{sec:datasets}
\subsection{Text Categorization Datasets}
We evaluated all approaches on $8$ supervised document datasets: 
($1$) BBCSPORT: BBC sports articles between 2004-2005, ($2$) TWITTER: a set of tweets labeled with sentiments ‘positive’, ‘negative’, or ‘neutral’ (the set is reduced due to the unavailability of some tweets), ($3$) RECIPE: a set of recipe procedure descriptions labeled by their region of origin, ($4$) OHSUMED: a collection of medical abstracts categorized by different cardiovascular disease groups (for computational efficiency we subsample the dataset, using the first 10 classes), ($5$) CLASSIC: sets of sentences from academic papers, labeled by publisher name, ($6$) REUTERS: a classic news dataset labeled by news topics (we use the 8-class version with train/test split as described in \cite{cachopo2007improving}, ($7$) AMAZON: a set of Amazon reviews which are labeled by category product in {books, dvd, electronics, kitchen} (as opposed to by sentiment), ($8$) 20NG: news articles classified into 20 different categories (we use the “bydate” train/test split by \cite{cachopo2007improving}).
We preprocess all datasets by removing all words in the SMART stop word list \citep{salton1971smart}.
Table~\ref{tab:datasets_text_categorization} shows statistics of the $8$ datasets that were used for the evaluation.
We obtained a distributed representation for each word from a publicly available set of pre-trained vectors\footnote{\url{https://code.google.com/archive/p/word2vec/}}.
For datasets that do not come with a predefined train/test split, we report the average accuracy over five $70/30$ train/test splits as well as the standard deviation.

\begin{table}[t]
\centering
\caption{Datasets used in text categorization experiments.}
\resizebox{0.9\columnwidth}{!}{
\begin{tabular}{lcccc}
\toprule
 & & & Unique & \\
 Dataset & $n$ & Voc & Words(avg) & $y$ \\
\midrule
BBCSPORT & 517 & 13243 & 117 & 5 \\
TWITTER & 2176 & 6344 & 9.9 & 3 \\
RECIPE & 3059 & 5708 & 48.5 & 15 \\
OHSUMED & 3999 & 31789 & 59.2 &10 \\
CLASSIC & 4965 & 24277 & 38.6 & 4\\
REUTERS & 5485 & 22425 & 37.1 & 8 \\
AMAZON & 5600 & 42063 & 45.0 & 4 \\
20NG & 11293 & 29671 & 72 & 20 \\
\bottomrule
\end{tabular}
}
\label{tab:datasets_text_categorization}
\end{table}

\subsection{Graph Classification Datasets}
We evaluated the proposed architecture on the following $5$ datasets: ($1$)  MUTAG, ($2$) PROTEINS, ($3$) IMDB-BINARY, ($4$) IMDB-MULTI, and ($5$) REDDIT-BINARY.

MUTAG contains mutagenic aromatic and heteroaromatic nitro compounds.
Each chemical compound is labeled according to whether or not it has mutagenic effect on the Gram-negative bacterium Salmonella typhimurium \citep{debnath1991structure}.
PROTEINS consists of proteins represented as graphs where vertices are secondary structure elements and there is an edge between two vertices if they are neighbors in the amino-acid sequence or in 3D space.
The task is to classify proteins into enzymes and non-enzymes \citep{borgwardt2005protein}.
IMDB-BINARY and IMDB-MULTI contain movie collaboration graphs.
The vertices of each graph represent actors/actresses and two vertices are connected by an edge if the corresponding actors/actresses appear in the same movie.
Each graph is the ego-network of an actor/actress, and the task is to predict which genre an ego-network belongs to \citep{yanardag2015deep}.
REDDIT-BINARY consists of online discussion threads represented as graphs.
Each vertex corresponds to a user, and two users are connected by an edge if one of them responded to at least one of the other's comments.
The task is to classify graphs into either communities \citep{yanardag2015deep}.
A summary of the datasets is given in Table~\ref{tab:datasets_graph_classification}.

\begin{table}[t]
\centering
\caption{Datasets used in graph classification.}
\resizebox{\columnwidth}{!} {
\begin{tabular}{lcccc}
\toprule
 Dataset & \#Graphs & $y$ & Nodes(avg) & Edges(avg) \\
\midrule
MUTAG & 188 & 2 & 17.93 & 19.79 \\
PROTEINS & 1113 & 2 & 39.06 & 72.82 \\
IMDB BINARY & 1000 & 2 & 19.77 & 96.53 \\
IMDB MULTI & 1500 & 3 & 13.00 & 65.94 \\
REDDIT BINARY & 2000 & 2 & 429.63 & 497.75 \\
\bottomrule
\end{tabular}
}
\label{tab:datasets_graph_classification}
\end{table}

\section{Experimental Evaluation}\label{sec:further_experiments}

\subsection{Sensitivity Analysis}
\begin{figure*}[t]
\centering
    \begin{minipage}{\columnwidth}
    \centering
    \resizebox{\columnwidth}{!}{%
    \includegraphics{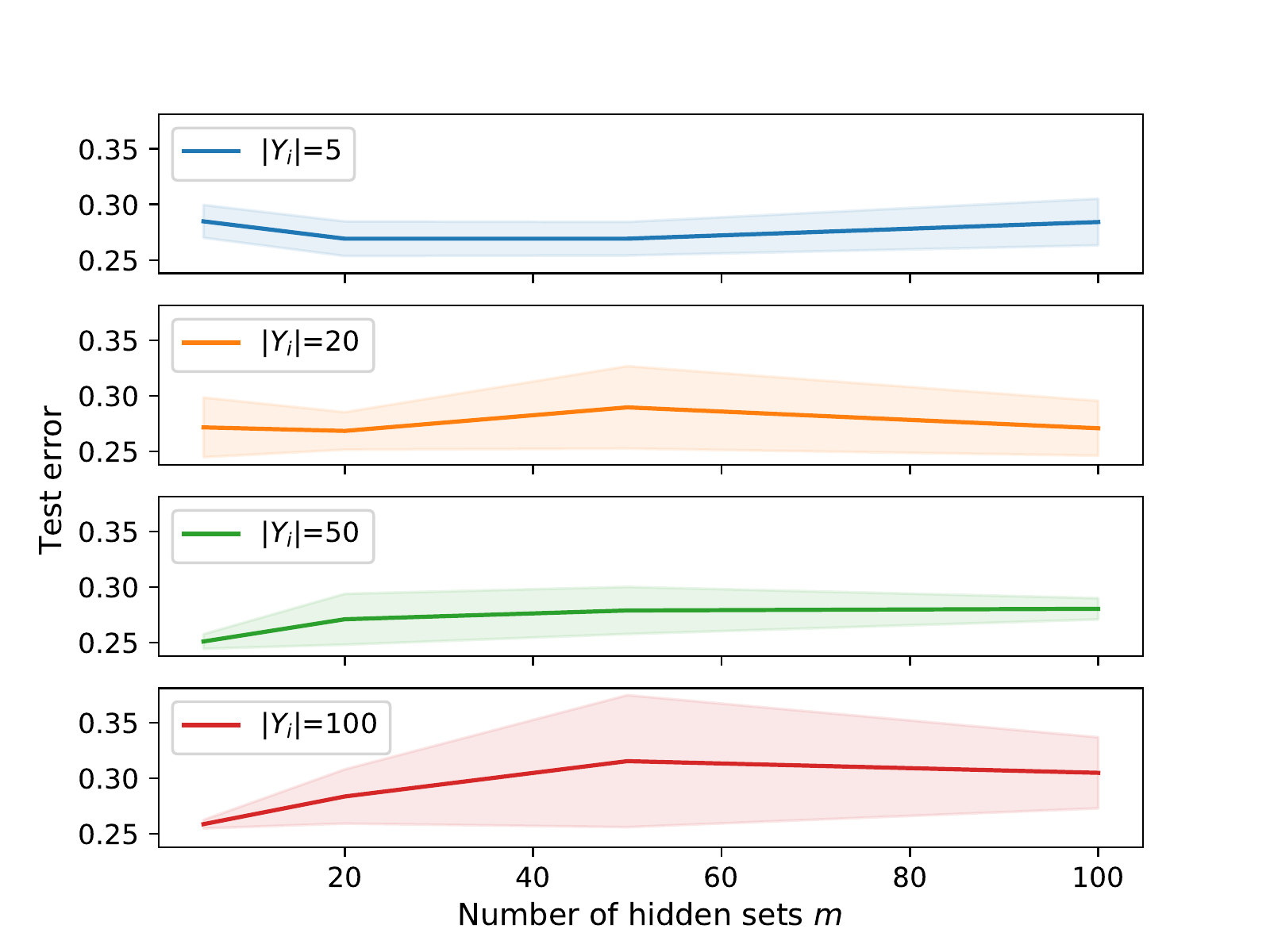}
    }
    %\caption{Different number of hidden sets.}
    %\label{fig:repset_sets}

\end{minipage}%
\begin{minipage}{\columnwidth}
    \centering
    \resizebox{\columnwidth}{!}{%
    \includegraphics{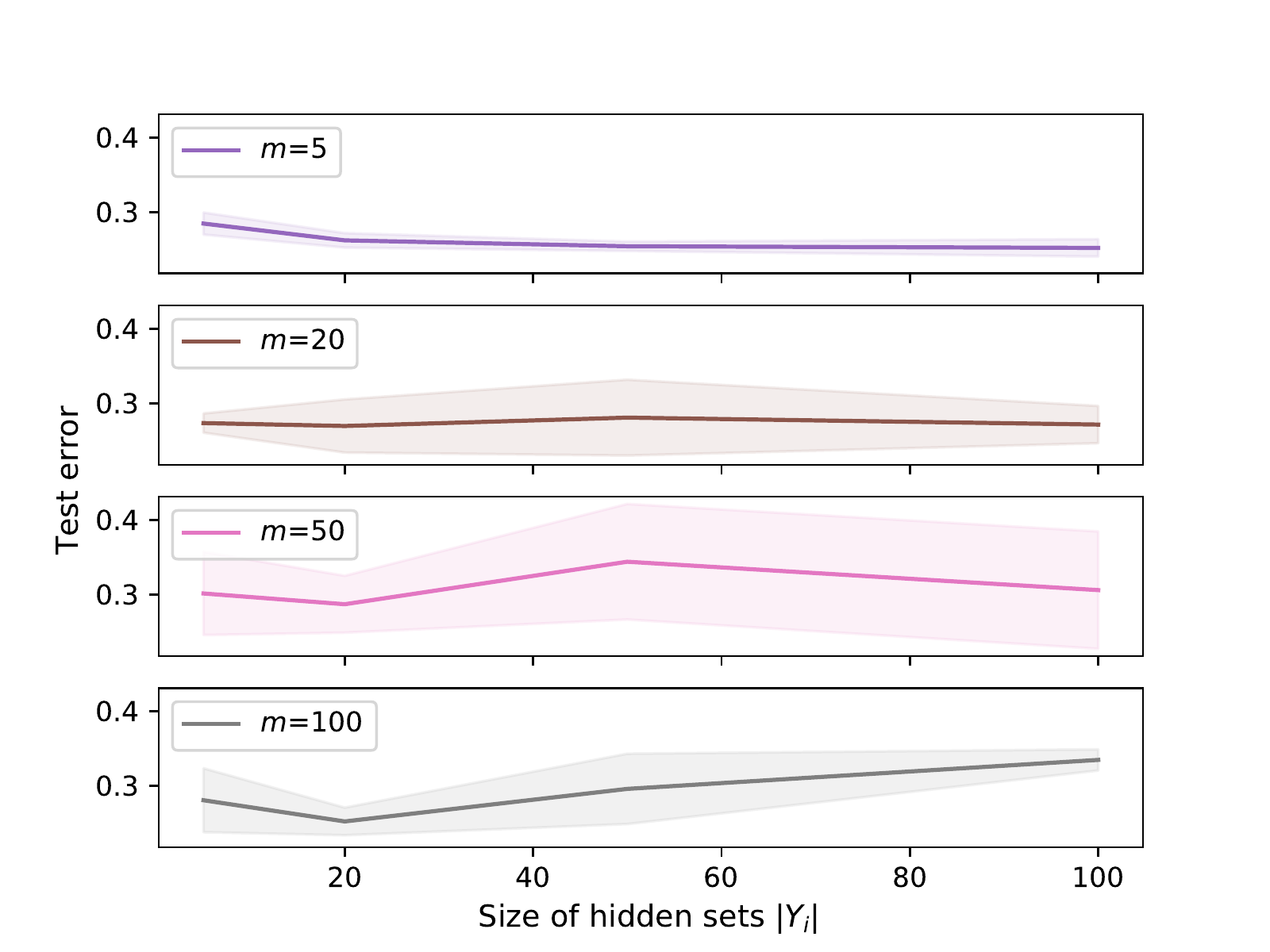}
    }
    %\caption{Different sizes of hidden sets.}
    %\label{fig:repset_size}
\end{minipage}
\caption{Average test error of the RepSet model with respect to the number of hidden sets $m$ (left) and the size of the hidden sets $|Y_i|$ (right) on the TWITTER dataset.}
\label{fig:twitter_parameter_sensitivity}
\end{figure*}

\begin{figure*}[t]
\centering
    \begin{minipage}{\columnwidth}
    \centering
    \resizebox{\columnwidth}{!}{%
    \includegraphics{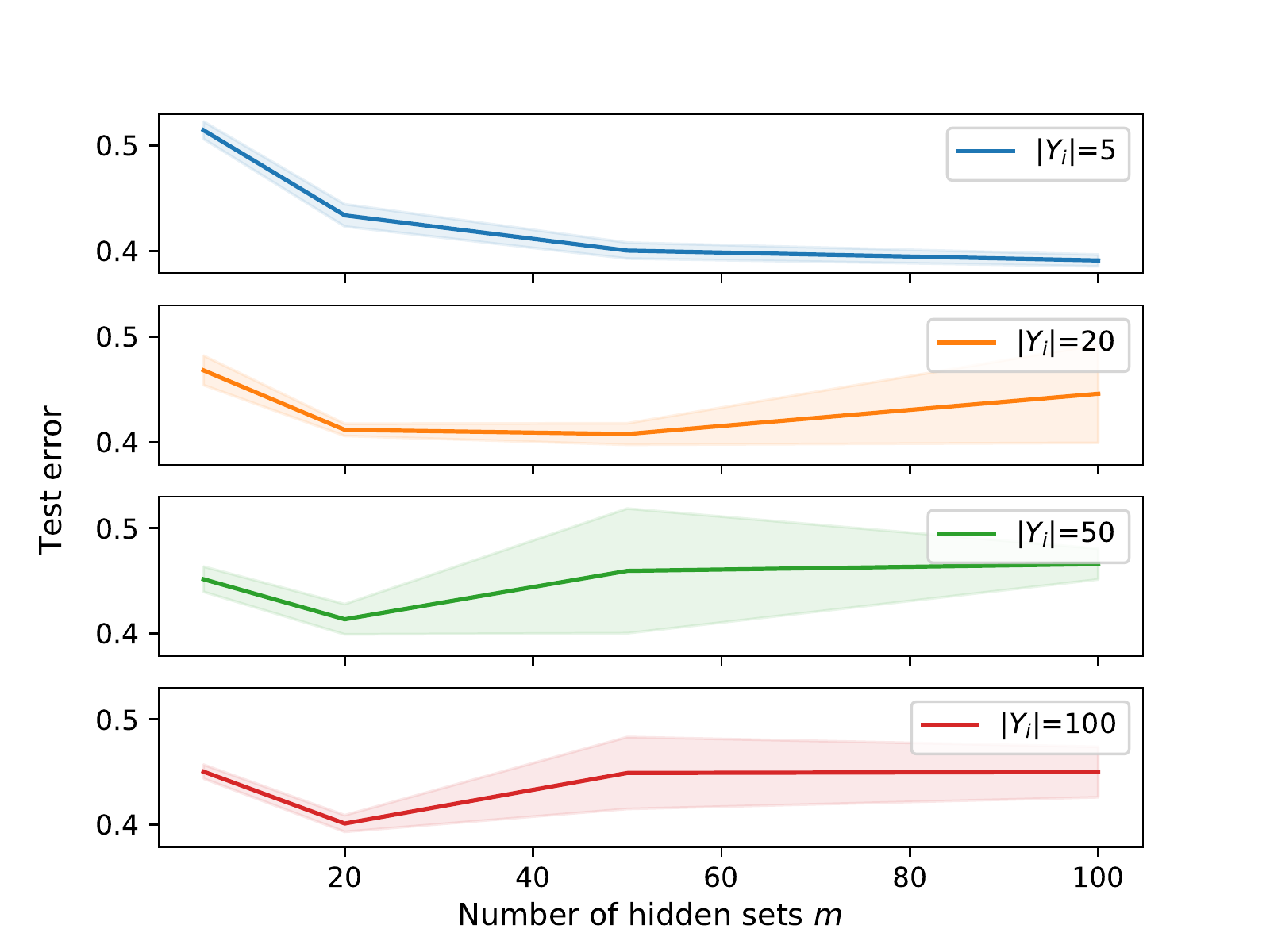}
    }
    %\caption{Different number of hidden sets.}
    %\label{fig:repset_sets}
\end{minipage}%
\begin{minipage}{\columnwidth}
    \centering
    \resizebox{\columnwidth}{!}{%
    \includegraphics{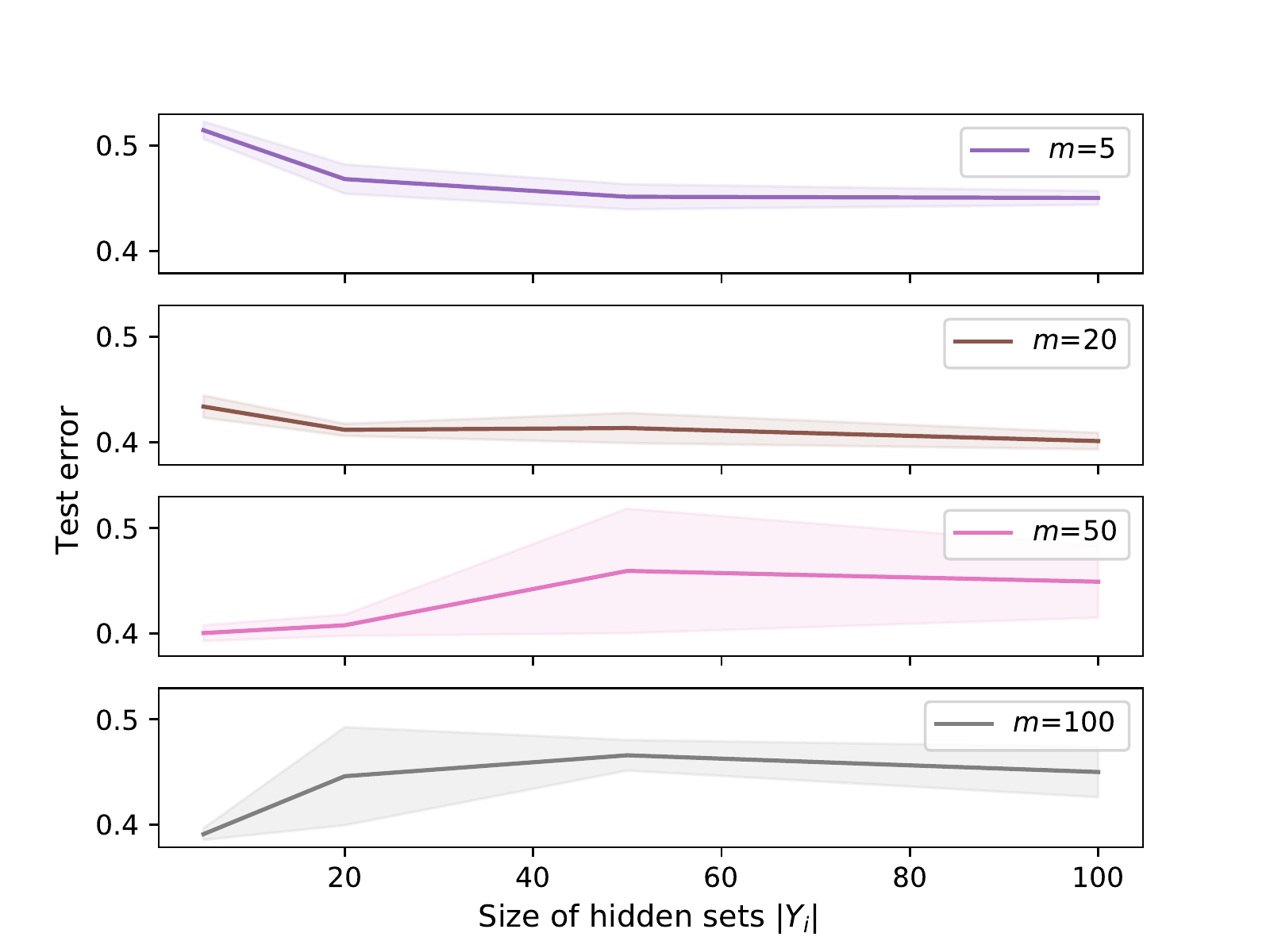}
    }
    %\caption{Different sizes of hidden sets.}
    %\label{fig:repset_size}
\end{minipage}
\caption{Average test error of the RepSet model with respect to the number of hidden sets $m$ (left) and the size of the hidden sets $|Y_i|$ (right) on the RECIPE dataset.}
\label{fig:recipe_parameter_sensitivity}
\end{figure*}

The proposed RepSet and ApproxRepSet models involve two main parameters:
($1$) the number of hidden sets $m$, and ($2$) the cardinalities of the hidden sets $|Y_i|, i=1,\ldots,m$.
We next investigate how these two parameters influence the performance of the RepSet model.
Specifically, in Figures~\ref{fig:twitter_parameter_sensitivity} and ~\ref{fig:recipe_parameter_sensitivity}, we examine how the different choices of these parameters affect the performance of RepSet on the TWITTER and RECIPE datasets, respectively. 
We measure the test error as a function of the two parameters.  
Note that each hidden set $Y_i$ can have a different cardinality compared to the other sets.
However, we set the cardinalities of all hidden sets to the same value.
We observe that on TWITTER, the number of hidden sets $m$ does not have a large impact on the performance, especially for small cardinalities of the hidden sets ($|Y_i| \leq 50$).
For most cardinalities, the test error is within $1\%$ to $3\%$ when varying this parameter.

Furthermore, in most cases, the best performance is attained when the number of hidden sets is small ($m \leq 20$).
Similar behavior is also observed for the second parameter on the TWITTER dataset.
For most values of $m$, the test error changes only slightly when varying the cardinalities of the hidden sets.
For $m \geq 50$, the model produces best results when the cardinalities of the hidden sets $|Y_i|$ are close to $20$.
On the other hand, for small values of $m$, the model yields good performance even when the cardinalities of the hidden sets $|Y_i|$ are large.
On the RECIPE dataset, both parameters have a higher impact on the performance of the RepSet model.
In general, small values of $m$ lead to higher test error than larger values of $m$.
For most values of $|Y_i|$, values of $m$ between $20$ and $50$ result in the lowest test error.
As regards the size of the hidden sets $|Y_i|$, there is no consistency in the obtained results.
Specifically, for small values of $m$ ($m \leq 20$), large cardinalities of the hidden sets result in better performace, while for large values of $m$ ($m \geq 50$), small cardinalities lead to smaller error.

\subsection{Synthetic Data}
\begin{figure}[H]
    \centering
    \includegraphics[width=\columnwidth]{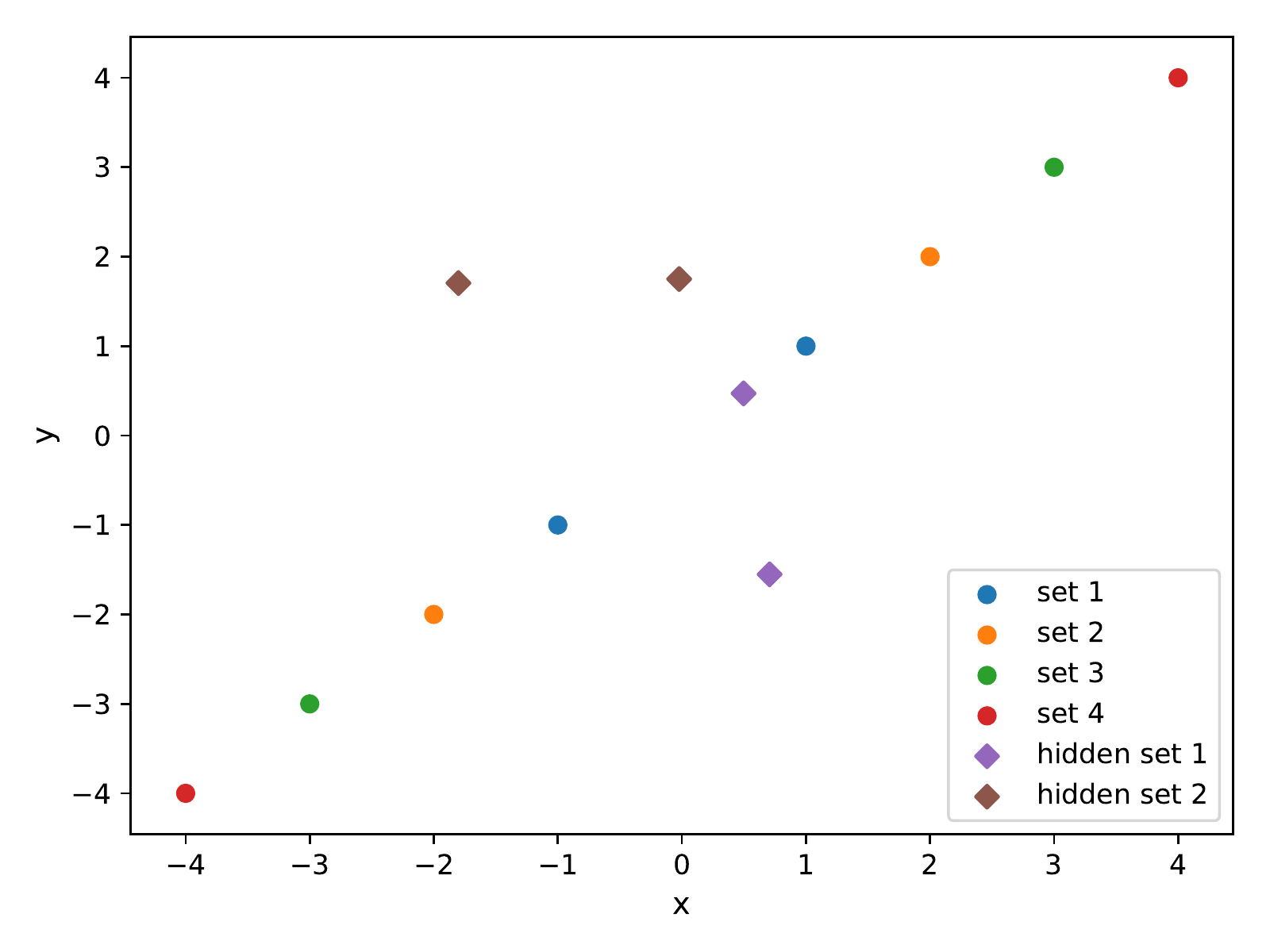}
    \caption{A very simple dataset consisting of $4$ examples (\ie sets). Each set contains a pair of $2$-dimensional vectors (circles). The diamonds indicate the ``hidden sets'' that the proposed model learned during training.}
    \label{fig:toy_example}
\end{figure}

We first demonstrate the proposed RepSet architecture on a very simple dataset.
The dataset consists of $4$ sets of $2$-dimensional vectors.
The cardinality of all sets is equal to $2$.
The elements of the $4$ sets are illustrated in Figure~\ref{fig:toy_example}. 
Although seemingly simple, this dataset may prove challenging for several algorithms that apply aggregation mechanisms to the elements of the sets since all $4$ sets have identical centroids while the sum of their elements is also the same for all of them.
To learn to classify these sets, we used an instance of the proposed model consisting of $2$ hidden sets of cardinality equal to $2$.
The model managed easily to discriminate between the $4$ sets and to achieve perfect accuracy.
A question that arises at this point is what kind of features the hidden sets of the model learn during training.
Hence, besides the input sets, Figure~\ref{fig:toy_example} also shows the vectors of the $2$ hidden sets.
The hidden sets learned very similar patterns, which indicates that less than $2$ hidden sets may be required.
In fact, we observed that even with one hidden set, the model can achieve perfect performance.

\end{document}